\documentclass[runningheads]{llncs}
\usepackage{graphicx}
\usepackage{comment}
\usepackage{amsmath,amssymb} 
\usepackage{color}
\usepackage[misc]{ifsym}

\usepackage{bm}
\usepackage[colorlinks,linkcolor=red]{hyperref}
\begin{document}
\pagestyle{headings}
\mainmatter
\def\ECCVSubNumber{3307}  

\title{Adversarial Semantic Data Augmentation for Human Pose Estimation} 


\titlerunning{ASDA for Human Pose Estimation}
%
\author{Yanrui Bin\inst{1}\orcidID{0000-0003-2845-3928} \and
Xuan Cao\inst{2} \and
Xinya Chen\inst{1}\orcidID{0000-0002-6537-4316} \and
Yanhao Ge\inst{2} \and
Ying Tai\inst{2} \and
Chengjie Wang\inst{2} \and
Jilin Li\inst{2} \and
Feiyue Huang\inst{2} \and
Changxin Gao\inst{1}\orcidID{0000-0003-2736-3920} \and
Nong Sang\inst{1}\orcidID{0000-0002-9167-1496}\thanks{Corresponding author.}}
%
\authorrunning{Y. Bin, X. Cao, X. Chen et al.}
%
\institute{Key Laboratory of Image Processing and Intelligent Control, School of Artificial Intelligence and Automation, Huazhong University of Science and Technology, Wuhan, China\\ \email{ \{yrbin, hust\_cxy, cgao, nsang\}@hust.edu.cn} \\
	 \and
Tencent Youtu Lab \\ \email{\{marscao, halege, yingtai, jasoncjwang, jerolinli, garyhuang\}@tencent.com}}
\maketitle

\begin{abstract}
Human pose estimation is the task of localizing body keypoints from still images. 
The state-of-the-art methods suffer from insufficient examples of challenging cases such as symmetric appearance, heavy occlusion and nearby person. 
To enlarge the amounts of challenging cases, previous methods augmented images by cropping and pasting image patches with weak semantics, which leads to unrealistic appearance and limited diversity.
We instead propose Semantic Data Augmentation (SDA), a method that augments images by pasting segmented body parts with various semantic granularity.
Furthermore, 
we propose Adversarial Semantic Data Augmentation (ASDA), which exploits a generative network to dynamiclly predict tailored pasting configuration. 
Given off-the-shelf pose estimation network as discriminator, the generator seeks the most confusing transformation to increase the loss of the discriminator while the discriminator takes the generated sample as input and learns from it.  
The whole pipeline is optimized in an adversarial manner.
State-of-the-art results are achieved on challenging benchmarks.
The code has been publicly available at \url{https://github.com/Binyr/ASDA}.

\keywords{Pose Estimation, Semantic Data Augmentation}
\end{abstract}

\section{Introduction}
Human Pose Estimation (HPE) is the task of localizing body keypoint from still images. It serves as a fundamental technique for numerous computer vision applications. Recently, deep convolutional neural networks (DCNN)~\cite{sun2019deep,ke2018multi,zhang2019human} have achieved drastic improvements on standard benchmark datasets.  However, as shown in Figure 1, they are still prone to fail in some challenging cases such as symmetric appearance, heavy occlusion, and nearby persons. 

The reason for the inferior performance of the DCNN-based methods in the challenging cases is that there exists an insufficient amount of examples that contain these challenging cases to train a deep network for accurate keypoint localization. 
However, obtaining the annotations of keypoint localization is costly. 

\begin{figure}[t!]
	\centering
	\includegraphics[width=1.0\columnwidth]{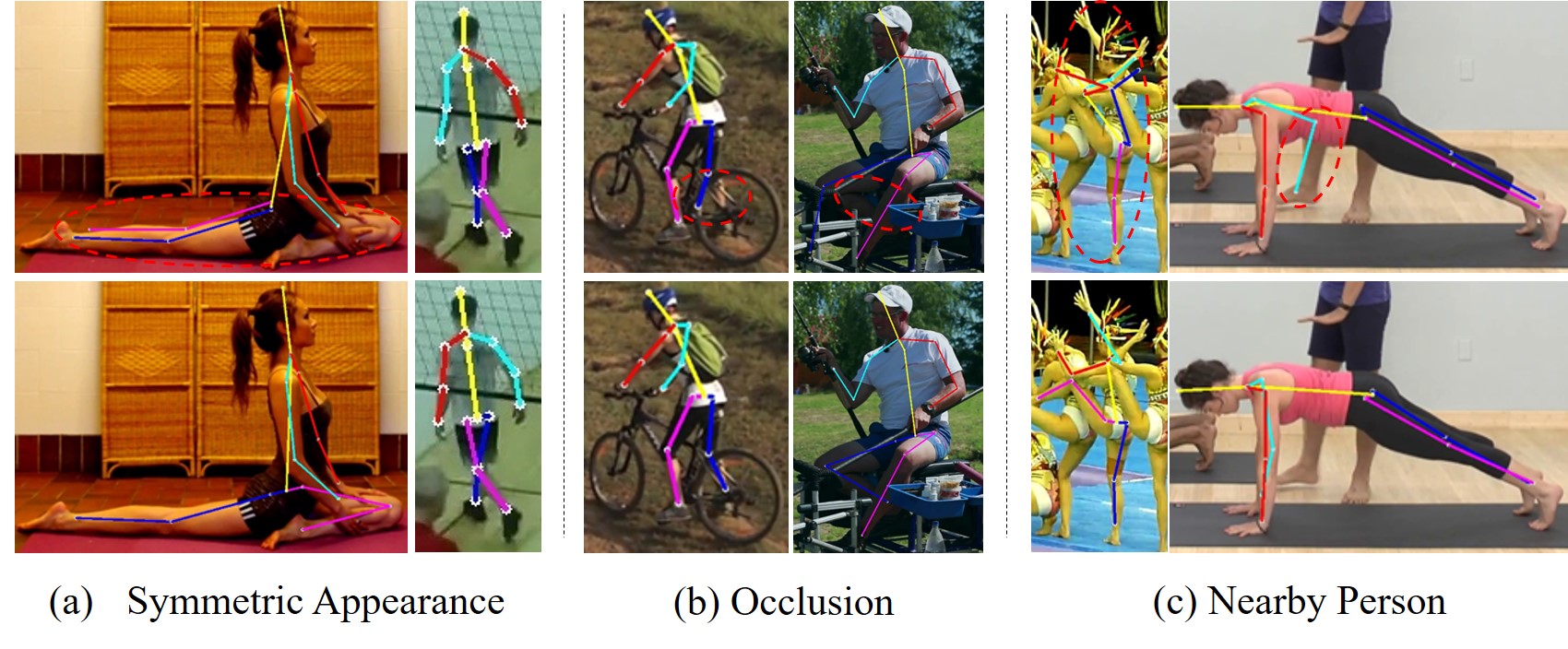}
	\caption{Pairs of pose predictions obtained by HRNet~\cite{sun2019deep} (top) and our approach (bottom) in the challenging cases. Incorrect predictions are highlighted by the red dotted circles. Note that image in Figure~\ref{fig:challeng_case}~(c)~\textit{ \{cols. 1\}} is an extremely challenging case so that few of the keyponts are correctly predicted by the original HRNet. After equipped with our ASDA (bottom), HRNet improve the robustness to the challenging cases.}
	\label{fig:challeng_case}
\end{figure}

One promising way to tackle this problem is data augmentation. Conventional data augmentation performs global image transformations (e.g., scaling, rotating, flipping or color jittering). Although it enhances the global translational invariance of the network and largely improves the generalizability, it contributes little to solving the challenging cases. Recently, Ke et al. \cite{ke2018multi} proposes keypoints masking training to force the network better recognize poses from difficult training samples.  They simulate the keypoint occlusion by copying a background patch and putting it onto a keypoint or simulate the multiple existing keypoints by copying a body keypoint patch and putting it onto a nearby background. However, this data augmentation method only brings marginal improvement. On the one hand, the used patch is cropped from the input image, resulting in a limited variance of the generated images. On the other hand, the cropped keypoint patch is surrounded by some background, which makes the generated image unrealistic.

In this paper, we propose a novel Adversarial Semantic Data Augmentation (ASDA) scheme. Human parsing is applied to the training images to get a large amount of pure body part patches. These body parts are organized, according to their semantic types, to build a semantic part pool. As the human body could be represented as a hierarchy of parts and subparts, we combine several subparts, according to the structure of the human body, to get body parts with various semantic granularity. For each input image, several parts will be randomly selected from the semantic part pool and properly pasted to the image. 

Further, randomly pasting parts to the image is still suboptimal. Without taking the difference between training image instances into account, it may generate ineffective examples that are too easy to boost the network. Moreover, it can hardly match the dynamic training status of the pose estimation network, since it is usually sampled from static distributions~\cite{peng2018jointly}. For instance, with the training of the network, it may gradually learn to associate occluded wrists while still have difficulty in distinguish similar appearance with legs.

Based on the above consideration, we parameterize the parts pasting process as an affine transformation matrix and exploit a generative network to online predict the transformation parameters. The generator seeks the most confusing transformation to increase the loss of the pose estimation network and consequently generates tailored training samples. The pose estimation network acts as a discriminator, which takes the tailored samples as input and tries to learn from it. By leveraging the spatial transformer network, the whole process is differentiable and trained in an adversarial manner.

Additionally, our Adversarial Semantic Data Augmentation is a universal solution that can be easily applied to different datasets and networks for human pose estimation.

In summary, the main contributions are three-fold:

$\bullet$ We design a novel Semantic Data Augmentation (SDA) which augments images by pasting segmented body parts of various semantic granularity to simulate examples that contain challenging cases.

$\bullet$ We propose to utilize a generative network to dynamically adjust the augmentation parameters of the SDA and produce tailored training samples against the pose estimation network, which largely elevates the performance of the SDA.

$\bullet$ We comprehensively evaluate our methods on various benchmark datasets and consistently outperforms the state-of-the-art methods.

\section{Related Work}
The advances of DCNN-based human pose estimation benefit from multiple factors. We compare our methods with literature from three most related aspects.

\subsection{Human Pose Estimation.}
Recently, pose estimation using DCNNs has shown superior performance. DeepPose~\cite{toshev2014deeppose} first applied deep neural networks to human pose estimation by directly regressing the 2D coordinates of keypoints from the input image. \cite{tompson2014joint} proposed a heatmap representation for each keypoint and largely improved the spatial generalization. Following the heatmap-based framework, various methods~\cite{wei2016convolutional,newell2016stacked,su2019cascade,tang2019does,sun2019deep,xiao2018simple,sun2019deep} focused on designing the structure of the network and indeed achieved significant improvement. 
However they still suffered from insufficient amounts of samples that contains challenging cases. In this work, standing on the shoulder of the well-designed network structure, we propose a universal data augmentation solution to further improve the performance of human pose estimation.

\subsection{Data Augmentation.}
Typical data augmentation~\cite{newell2016stacked,chen2017adversarial,xiao2018simple,sun2019deep} mainly performed global spatial transformation like scaling, rotating and flipping \textit{etc}. These common data augmentation schemes helped the network to resist the global image deformation but fail to improve the immunity to the challenging cases.
Recently, some novel data augmentations were proposed. 
PoseRefiner~\cite{fieraru2018learning} transformed the keypoint annotations to mimic the most common failure cases of human pose estimators, so that the proposed refiner network could be trained well.
MSR-net~\cite{ke2018multi} introduced keypoint-masking which cropped and pasted patches from the input image to simulate challenging cases. Different from the existing data augmentation strategies, we propose a novel semantic data augmentation scheme which takes advantage of the human semantic segmentation to obtain the pure segmented body parts rather than noisy image patches. Furthermore, we compose the related parts to form a set of new parts with higher semantic granularity.

\subsection{Adversarial Learning.}
Inspired by the minimax mechanism of Generative Adversarial Networks (GANs) \cite{goodfellow2014generative}, some literature~\cite{Chu_2019} generated hard training samples in an adversarial way. Semantic Jitter~\cite{yu2017semantic} proposed to overcome the sparsity of supervision problem via synthetically generated images. A-Fast-RCNN~\cite{wang2017fast} used GANs to generate deformations for object detection. 
Recently, GANs were introduced into human pose estimation. Such as Adversarial PoseNet~\cite{chen2017adversarial} designed discriminators to distinguish the real poses from the fake ones. Jointly Optimize~\cite{peng2018jointly} designed an augmentation network that competed against a target network by generating “hard” augmentation operations.
In this paper, we designed a generative network to adjust the semantic data augmentation then to produce challenging training data. The generative network takes the difference between training instances into consideration, and produce tailored training samples for the pose estimation network. Hard mining, as an alternative strategy to feed challenging training data to network, is totally different from ours. Hard mining can only "select" rather than "produce" challenging samples, which essentially limits its improvement of accuracy on challenging cases.

\begin{figure}[t]
	\centering
	\includegraphics[width=1.0\columnwidth]{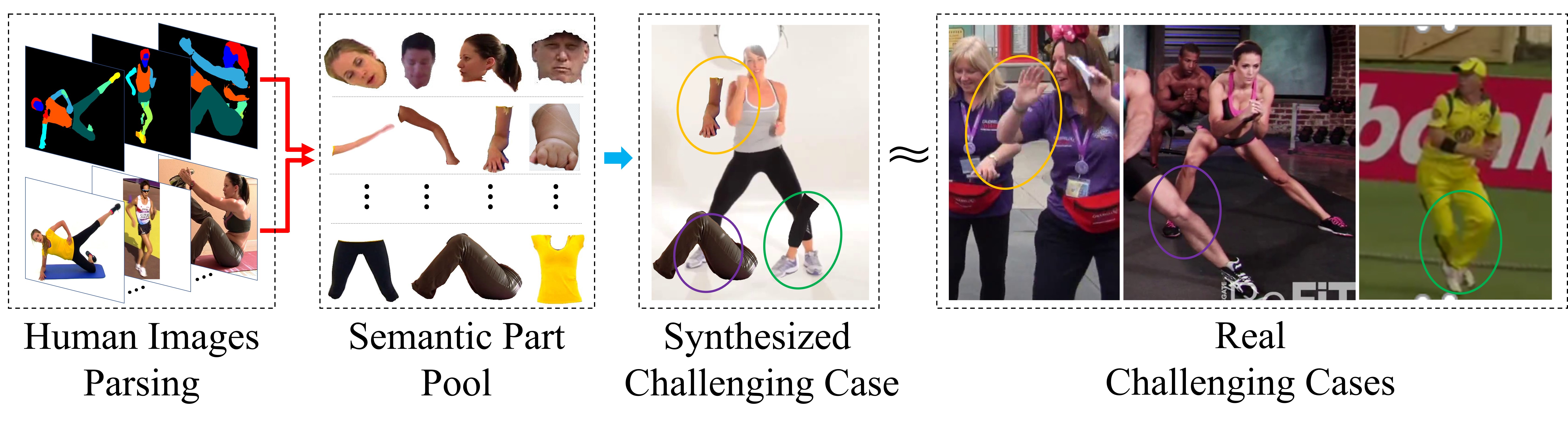}
	\caption{Illustration of Semantic Data Augmentation (SDA). We first apply human parsing on training images and get a large amount of segmented body parts.  The segmented body parts are organized, according to  their semantics, to build semantic part pool. For each training image, several part patches will be randomly sampled and properly placed on the image to synthesize the real challenging cases such as symmetric appearance (green circle), occlusion (perple circle) and nearby person (yellow circle).}
	\label{fig:SDA}
\end{figure}

\section{Methodology}
\subsection{Semantic Data Augmentation}\label{subsec:SDA}
\textbf{Building Semantic Part Pool.} For common human pose estimation schemes \cite{newell2016stacked,xiao2018simple,tang2018deeply,sun2019deep}, data augmentations such as global scaling, rotation, flipping are usually applied, which bring the global translational invariances to the network and largely improves the generalizability. 

However, the remained problem of pose estimation task is the challenging cases, e.g., symmetric appearance, heavy occlusion, and nearby person, where the global spatial transformation helps little. 
In contrast to the global spatial transformations, local pixel patch manipulation provide more degrees of freedom to augment image and is able to synthesize the challenging case realistically. 

A human image is assembled by semantic part patches, such as arm, leg, shoe, trousers and so on. Inspired by these semantic cues, we can synthesize plentiful human image instances by elaborately combining these local part patches.
Here, we propose a novel augmentation scheme, as shown in Figure~\ref{fig:SDA}. By firstly segmenting all human images through the human parsing method~\cite{liu2018devil},  then we can build a data pool $\mathbb{D}_{part}$ filled with various semantic body part patches. We follow the definition of LIP dataset~\cite{gong2017look} and segment the human image into $\hat{N}=26$ part patches. Finally, the body part patches from the data pool can be properly mounted on the current person's body to synthesize challenging cases.

As human parsing aims to analyze every detail region of a person as well as different categories of clothes, LIP defines 6 body parts and 13 clothes categories in fine semantic granularity. However, body parts of various semantic granularity will appear in images of real-world scenarios with complex multi-person activities. For the above considerations, we combine some of the parts (e.g., left shoe and left leg) to form a set of new parts with higher semantic granularity and then add them to our part pool. 
After the cutting step, we filter out scattered segments, segments with the area below $35^2$ and segments with low semantics.

\textbf{Augmentation Parameter Formulation.} Given a semantic part patch $I_p$ and a training image $I_o$, the placement of this semantic part can be defined by the affine transformation matrix 
\begin{equation}
\bm{H} = \begin{bmatrix} s\cos r & s\sin r & t_x \\ 
-s\sin r & s\cos r & t_y \\
0 & 0 & 1\end{bmatrix} \,,
\label{eq:matrix}
\end{equation}
where $s$ denotes the scale of the part patch, $r$ denotes the rotation, and $t_x, t_y$ is the translation in horizontal and vertical direction respectively. Thus the placement of the part patch $I_p$  can be uniquely determined by a 4D tuple $\theta(s,r,t_x,t_y)$.

The scale of the part patch will be aligned with the target person in advance according to the human bounding box. Initially, the part patch could be pasted in the center of the training image without rotation. In other words, the tuple $(1,0,0,0)$ is served as our original paste configuration.

\textbf{Random Semantic Augmentation.} With 4D augmentation parameters defined in Equation~\ref{eq:matrix}, a straight augmentation method can be realized by sampling a 4D tuple augmentation parameter from a uniform distribution in the neighborhood space of $(1,0,0,0)$. $N$ different body parts will be pasted to the target person. The value of $N$ is set manually as a hyper-parameter. Sensitivity Analysis of $N$ is detailed in Section~\ref{exp:ablation}.
%

\begin{figure}[t]
	\centering
	\includegraphics[width=1.0\columnwidth]{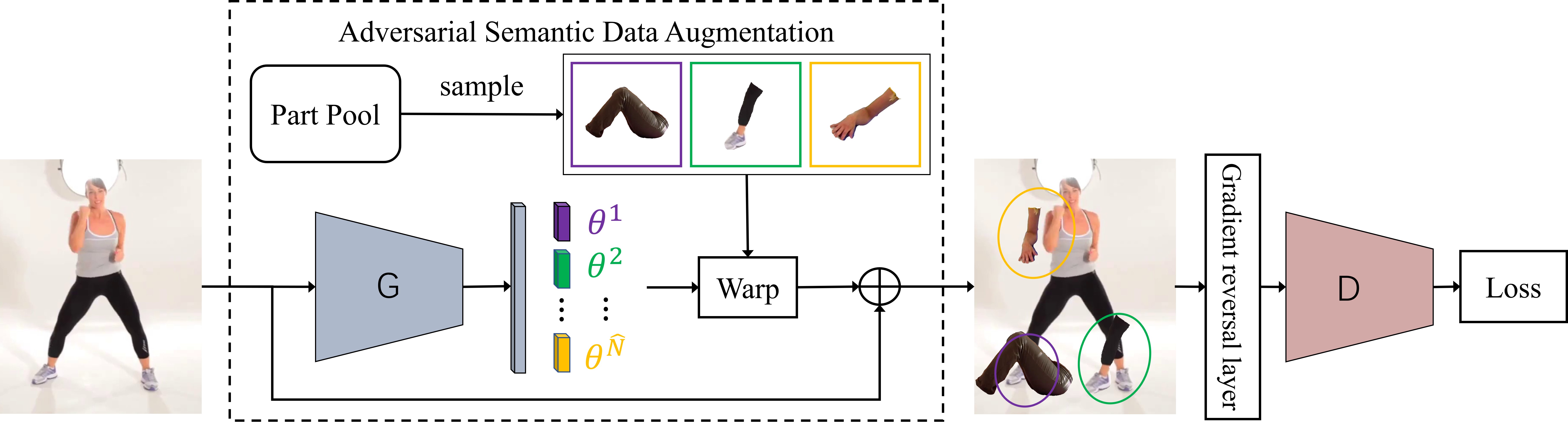}
	\caption{ Overview of our approach. The input image is fed to the generator $\mathcal{G}$ to obtain $\hat{N}$ groups of tailored augmentation parameters which are used to warp the randomly selected semantic part patches. Each group parameters is used to warp the patch of the specific part type. $\mathcal{G}$ seeks the most confusing transformation to increase the loss of the pose estimation network and consequently generates tailored training samples. The pose estimation network acts as a discriminator $\mathcal{D}$, which takes the tailored sample as input and tries to learn from it. The whole pipeline is optimized in an adversarial manner. }
	\label{fig:pipeline}
\end{figure}

\subsection{Adversarial Learning} \label{sec:adversarial_learning}
Our goal is to generate the confusing transformation to improve the performance of pose estimation networks.  However, the augmentation parameters of SDA are sampled from the neighborhood of $(1,0,0,0)$. On the one hand, the confusing transformation naturally varies with different training instances and different part types. On the other hand, random sampling augmentation parameters from the static distribution can hardly perceive the dynamic training status. Thus it is prone to generate ineffective training samples which are so easy that it may not bring positive or even put negative effect on network training.

To overcome such issues, we propose to leverage Spatial Transformer Network (STN) to manipulate semantic parts within the network and optimize it in an adversarial manner. The main idea is to utilize an STN as the generator, which seeks the most confusing transformation to increase the pose estimation network loss. On the other hand, the pose estimation network acts as a discriminator, which tries to learn from the tailored semantic augmentation.

\textbf{Generate Tailored Samples.} 
The core module of our method is an STN, which takes the target person image as input and predicts $\hat{N}$ groups transformation parameters, each of which is used to transform the randomly selected semantic body parts of the specific part type.  In our experiments, we find that allowing the network to predict the scale $s$ of the part would collapse the training. It would easily predict a large scale, so that the part completely covers the target person in the training images. Thus, we randomly sample the scale $s$ from the neighboring space of $1.0$ and the generative network is mainly responsible for predicting the $(r,t_x,t_y)$. The affine transformation matrix is generated as defined in Equation~\ref{eq:matrix}. 

Each pixel in the transformed image is computed by applying a sampling kernel centered at a particular location in the original image. Mathematically, the pointwise transformation is shown in eq. (\ref{eq:trans}).
\begin{equation}
\label{eq:trans}
\begin{pmatrix}x^s_i \\ y^s_i \\ 1 \end{pmatrix} = 
\bm{H}\begin{pmatrix}x^t_i \\ y^t_i \\ 1\end{pmatrix} \,,
\end{equation}
where $(x^s_i, y^s_i)$ and $(x^t_i, y^t_i)$ denote the coordinates of the i-th pixel in the original and transformed image respectively. The transformed parts thus can be pasted to the target person image in the order they were sampled.

It is the not the first time to determine the augmentation parameters through a network. 
Xi Peng et al~\cite{peng2018jointly} jointly optimizes the conventional data augmentation (i.e., global scaling, rotating and feature erasing.) and network training to enhance the global transformation invariance of the network.
Our contributions are quite different with~\cite{peng2018jointly}. We design a novel SDA which augments images by pasting segmented body parts of various semantic granularity to simulate examples that contain challenging cases. Then we further propose ASDA that utilize a generative network to dynamically adjust the augmentation parameters of the SDA and produce tailored training samples for the pose estimation network.
%

\textbf{Joint Training.} 
As shown in the Figure~\ref{fig:pipeline}, the networks training follow the pipeline of training standard GANs \cite{goodfellow2014generative}. Generative network acting as generator $\mathcal{G}$ try to produce challenging cases. Meanwhile, the pose estimation network acting as a discriminator $\mathcal{D}$ try to learn from the generated training samples.

The discriminator is supervised by ground-truth heatmaps and try to decrease the loss $\mathcal{L_D}$ which is formulated as eq. (\ref{eq_lossD}). On the contrary, the generator try to increase the loss $\mathcal{L_D}$. So the loss  for generator is simply set as negative discriminator loss as formulated in eq. (\ref{eq_lossG}).
\begin{equation}
I_{aug} = \mathcal{F}_{aff} (\mathcal{G}(I_o), \{ I_p \})\,,
\end{equation}
\begin{equation}
\label{eq_lossD}
\mathcal{L_D} = \left \| \mathcal{D} (I_{aug}) - H_{gt} \right\|_{\ell_2}\,, 
\end{equation}
\begin{equation}
\label{eq_lossG}
\mathcal{L_G} = -\mathcal{L_D}\,,
\end{equation}
where $I_o$ is the original training image, $\{I_p\}$ is a set of randomly sampled part patches, $\mathcal{F}_{aff}(\cdot , \cdot)$ denotes the affine transformation function and $H_{gt}$ denote ground-truth heatmap. 
The network weights of $\mathcal{G}$ and $\mathcal{D}$ are updated alternately.

\section{Experiments}
\subsection{Datasets and Evaluation Protocols}
We conduct experiments on three representative benchmark datasets, \textit{i.e.} extended Leeds Sports Poses (LSP) dataset~\cite{johnson2010clustered}, MPII human pose dataset~\cite{andriluka20142d} and MS COCO dataset~\cite{lin2014microsoft}. 

\textbf{LSP Dataset.} The extended LSP dataset consists of 11k training images and 1k testing images of mostly sports people. Standard Percentage of Correct Keypoints (PCK) metric is used for evaluation. It reports the percentage of keypoint that fall into a normalized distance of the ground-truth, where the torso size is used as the normalized distance. 

\textbf{MPII Dataset.} The MPII dataset includes around 25k images containing over 40k people with annotated body keypoint (28k training and 11k testing). Following \cite{newell2016stacked}, 3k samples are taken as a validation set to tune the hyper-parameters. PCK is also utilized to evaluate MPII, but distance is normalized by head size. MPII evaluation metric is referred to PCKh. 

\textbf{COCO Dataset.} The COCO dataset involves multi-person pose estimation task which requires simultaneously detecting people and localizing their key points. The COCO training dataset (train2017) includes 57k images and validation dataset (val2017) includes 5000 images. The COCO evaluation defines the object keypoint similarity (OKS) which plays the same role as the IoU. %

\subsection{Implementation Details} 
Both generator $\mathcal{G}$ and discriminator $\mathcal{D}$ are the off-the-shelf networks. For generator, the ResNet-18 is utilized to regress $(3 \times \hat{N})$ parameters, where $\hat{N}$ is the class number of the human parsing. For discriminator, we adopt HRNet~\cite{sun2019deep}.

During building the semantic part pool, in order to avoid the inference of different human parsing algorithms, we obtain body parts from LIP dataset~\cite{gong2017look}. Beside our semantic data augmentation, we keep original data augmentation as adopted in HRNet, including global random flip, rotation and scale.

Network training is implemented on the open-platform PyTorch. For training details, we employ Adam~\cite{kingma2014adam} with a learning rate 0.001 as the optimizer of both generator and discriminator network. We drop the learning rate by a factor of 10 at the 170-th and 200-th epochs. Training ends at 210 epochs. The HRNet is initialized with weight of pre-trained model on public-released ImageNet~\cite{imagenet}.

\textbf{MPII.} For both MPII training and testing set, body scale and center are provided. We first utilize these value to crop the image around the target person and resized to $256\times256$ or $384\times384$. Data augmentation includes random flip, random rotation $(-30^{\circ}, 30^{\circ})$ and random scale $(0.75, 1.25)$. 

\textbf{LSP.} For LSP training set, we crop image by estimating the body scale and position according to keypoint positions. The data augmentation strategy are the same to MPII. For the LSP testing set, we perform similar cropping and resizing, but simply use the image center as the body position, and estimate the body scale by the image size following~\cite{yang2017learning}. We follow previous methods~\cite{wei2016convolutional,yang2017learning} to train our model by adding the MPII training set to the extended LSP training set with person-centric annotations.  For both MPII and LSP, testing is conducted on six-scale image pyramids (0.8, 0.9, 1.0, 1.1, 1.2 1.3).

\textbf{COCO.} For COCO training set, each ground-truth human box is extended to fixed aspect ratio, e.g., height : width = 4 : 3  and enlarged to contain more context by a rescale factor 1.25. Then the resulting box is cropped from image without distorting image aspect ratio and resized to a fixed resolution.  The default resolution is 256 : 192. We apply random flip, random rotation $(-40^{\circ}, 40^{\circ})$ and random scale $(0.7, 1.3)$. For COCO testing set, we utilized the predicted bounding box released by Li et al~\cite{li2019rethinking}. We also predict the pose of the corresponding flipped image and average the heat maps to get the final prediction. 

\subsection{Quantitative Results}
We report the performance of our methods on the three benchmark datasets following the public evaluation protocols. We adopt the HRNet as the backbone network. "W32" and "W48" represent the channel dimentions of the high-resolution subnetworks in last three stages of HRNet, respectively. "s7" indicates the we expand the HRNet to 7 stages by repeating the last stage of the original HRNet.

\textbf{Results on LSP.} Table~\ref{table:lsp_test} presents the PCK@0.2 scores on LSP test set. Our method outperforms the state-of-the-art methods especially on some challenging keypoints, e.g., wrist, knee and ankle, we have 0.8\%, 1.0\% and 1.0\% improvements respectively.

\begin{table}[h]
	\centering
	\tabcolsep=2.0pt
	\caption{Comparisons on the LSP test set (PCK@0.2).}
	\centering
	\begin{tabular}{l|ccccccc|cc}
		\hline
		Method& Hea. &	Sho. &	Elb. & Wri.	& Hip.	& Kne.	& Ank. & Total \\
		\hline
		Insafutdinov et al., 2016~\cite{insafutdinov2016deepercut} & 97.4 & 92.7 & 87.5 & 84.4 & 91.5 & 89.9 & 87.2 & 90.1 \\
		Wei et al., 2016~\cite{wei2016convolutional} & $97.8$ & $92.5$ & $87.0$ & $83.9$ & $91.5$ & $90.8$ & $89.9$ & $90.5$ \\
		Bulat et al., 2016~\cite{bulat2016human} & 97.2 & 92.1 & 88.1 & 85.2 & 92.2 & 91.4 & 88.7 & 90.7 \\
		Chu et al., 2017~\cite{chu2017multi} & 98.1 & 93.7 & 89.3 & 86.9 & 93.4 & 94.0 & 92.5 & 92.6 \\
		Chen et al., 2017~\cite{chen2017adversarial} & 98.5 & 94.0 & 89.8 & 87.5 & 93.9 & 94.1 & 93.0 & 93.1 \\
		Yang et al., 2017~\cite{yang2017learning} & 98.3  & 94.5  & 92.2  & 88.9  & 94.4  & 95.0 & 93.7 & 93.9 \\
		Zhang et al., 2019~\cite{zhang2019human} & $98.4$ & $94.8$ & $92.0$ & $89.4$ & $94.4$ & $94.8$ & $93.8$ & $94.0$ \\
		\hline
		Ours-W32 & $\textbf{98.8}$  & $\textbf{95.2}$  & $\textbf{92.5}$  & $\textbf{90.2}$  & $\textbf{94.7}$  & $\textbf{95.8}$ & $\textbf{94.8}$ & $\textbf{94.6}$ \\	
		\hline
	\end{tabular}
	\label{table:lsp_test}
\end{table}

%
\textbf{Results on MPII.} The performance of our methods on MPII test set is shown in Table~\ref{table:mpii_test}. We can observe that Ours-W48-s7 achieves 94.1\% PCKh@0.5, which is the new state-of-the-art result. In particular, Ours-W48-s7 achieves 0.5\%, 0.5\% and 0.7\% improvements on wrist, knee and ankle which are considered as the most challenging keypoints.

\begin{table}[h]
	\centering
	\tabcolsep=2.0pt    
	\caption{Comparisons on the MPII test set (PCKh@0.5).}
	\begin{tabular}{l|ccccccc|c}
		\hline
		Method& Hea. &	Sho. &	Elb. & Wri.	& Hip.	& Kne.	& Ank. & Total \\
		\hline
		Wei et al., 2016~\cite{wei2016convolutional} & 97.8  & 95.0  & 88.7  & 84.0  & 88.4  & 82.8 & 79.4 & 88.5 \\
		Bulat et al., 2016~\cite{bulat2016human} & 97.9  & 95.1  & 89.9  & 85.3  & 89.4  & 85.7 & 81.7 & 89.7 \\
		Newell et al., 2016~\cite{newell2016stacked} & 98.2  & 96.3  & 91.2  & 87.1  & 90.1  & 87.4 & 83.6 & 90.9 \\
		Ning et al., 2018~\cite{ning2018knowledge}& 98.1  & 96.3  & 92.2  & 87.8  & 90.6  & 87.6 & 82.7 & 91.2 \\
		Chu et al., 2017~\cite{chu2017multi} & 98.5  & 96.3  & 91.9  & 88.1  & 90.6  & 88.0 & 85.0 & 91.5 \\
		Chen et al., 2017~\cite{chen2017adversarial} & 98.1  & 96.5  & 92.5  & 88.5  & 90.2  & 89.6 & 86.0 & 91.9 \\
		Yang et al., 2017~\cite{yang2017learning} & 98.5  & 96.7  & 92.5  & 88.7  & 91.1  & 88.6 & 86.0 & 92.0 \\
		Xiao et al., 2018~\cite{xiao2018simple} & 98.5  & 96.6  & 91.9  & 87.6  & 91.1  & 88.1 & 84.1 & 91.5 \\
		Ke et al., 2018~\cite{ke2018multi} & 98.5  & 96.8  & 92.7  & 88.4  & 90.6  & 89.4 & 86.3 & 92.1 \\
		Nie et al., 2018~\cite{nie2018human} & 98.6  & 96.9  & 93.0  & 89.1  & 91.7  & 89.0 & 86.2 & 92.4 \\ 
		Tang et al., 2018~\cite{tang2018deeply} & 98.4  & 96.9  & 92.6  & 88.7  & 91.8  & 89.4 & 86.2 & 92.3 \\
		Sun et al., 2019~\cite{sun2019deep} & 98.6  & 96.9  & 92.8  & 89.0 & 91.5  & 89.0 & 85.7 & 92.3 \\
		Zhang et al., 2019~\cite{zhang2019human} & 98.6 & 97.0 & 92.8 & 88.8 & 91.7 & 89.8 & 86.6 & 92.5 \\
		Su et al., 2019~\cite{su2019cascade}*& $98.7$  & $97.5$  & $94.3$  & $90.7$  & $\textbf{93.4}$  & $92.2$ & $88.4$ & $93.9$ \\
		\hline
		Ours-W48-s7* & $\textbf{98.9}$  & $\textbf{97.6}$  & $\textbf{94.6}$  & $\textbf{91.2}$  & $93.1$  & $\textbf{92.7}$ & $\textbf{89.1}$ & $\textbf{94.1}$ \\
		\hline
		\multicolumn{9}{l}{ "*" indicates the network take image size $384 \times 384$ as input.}
	\end{tabular}
	\label{table:mpii_test}
\end{table}

\textbf{Results on COCO.} Table~\ref{table:COCOval} compares our methods with classic and SOTA methods on COCO val2017 dataset. All the methods use standard top-down paradigm which sequentially performs human detection and single-person pose estimation. Our model outperforms SIM~\cite{xiao2018simple} and HRNet~\cite{sun2019deep} by 4.8\% and 0.8\% for input size $256\times192$ respectively. When input size is $384\times288$, our model achieve better AP than SIM~\cite{xiao2018simple} and HRNet~\cite{sun2019deep} by 4.5\% and 0.9\%.

\begin{table}[h]
	\centering
	\caption{Comparison with SOTA methods on COCO val2017 dataset. Their results are cited from Chen et al.~\cite{chen2018cascaded} and Sun et al.~\cite{sun2019deep}.}
	\resizebox{\textwidth}{!}{
	\begin{tabular}{l|c|c|c|c|cccccc}
			\hline
			Method & Backbone & Input Size & Params & GFLOPs & ${\rm AP}$ & ${\rm AP^{50}}$ & ${\rm AP^{75}}$ & ${\rm AP^M}$ & ${\rm AP^L}$ & ${\rm AR}$\\
			\hline
			Hourglass~\cite{newell2016stacked} & HG-8stage & $256\times192$ & 25.1M & 14.3 & 66.9 & - & - & - & - & - \\
			CPN~\cite{chen2018cascaded} & ResNet-50 & $256\times192$ &  27.0M & 6.20 & 69.4 & - & - & - & - & - \\
			CPN~\cite{chen2018cascaded} & ResNet-50 & $384\times288$ &  27.0M & 13.9 & 71.6 & - & - & - & - & - \\
			SIM~\cite{xiao2018simple} & ResNet-50 & $256\times192$ & 34.0M&8.9 & 70.4 & 88.6 & 78.3 & 67.1 & 77.2 & 76.3 \\
			SIM~\cite{xiao2018simple} & ResNet-50 & $384\times288$ &34.0M&20.0 & 72.2 & 89.3 & 78.9 & 68.1 & 79.7 & 77.6 \\
			HRNet~\cite{sun2019deep} & HRNet-W32 & $256\times192$ & 28.5M & 7.10 & 74.4 & 90.5 & 81.9 & 70.8 & 81.0 & 79.8 \\
			HRNet~\cite{sun2019deep} & HRNet-W32 & $384\times288$ & 28.5M & 16.0 & 75.8 & 90.6 & 82.7 & 71.9 & 82.8 & 81.0 \\
			\hline
			Ours & HRNet-W32 & $256\times192$ & 28.5M& 7.10 & $\bm{75.2}$ & $\bm{91.0}$ & $\bm{82.4}$ & $\bm{72.2}$ & $\bm{81.3}$ & $\bm{80.4}$ \\
			Ours & HRNet-W32 & $384\times288$ & 28.5M& 16.0 & $\bm{76.7}$ & $\bm{91.2}$ & $\bm{83.5}$ & $\bm{73.2}$ & $\bm{83.4}$ & $\bm{81.5}$ \\
			\hline
	\end{tabular}
	}
	\label{table:COCOval}
\end{table}

\begin{figure}[t!]
	\centering
	\includegraphics[width=1.0\columnwidth]{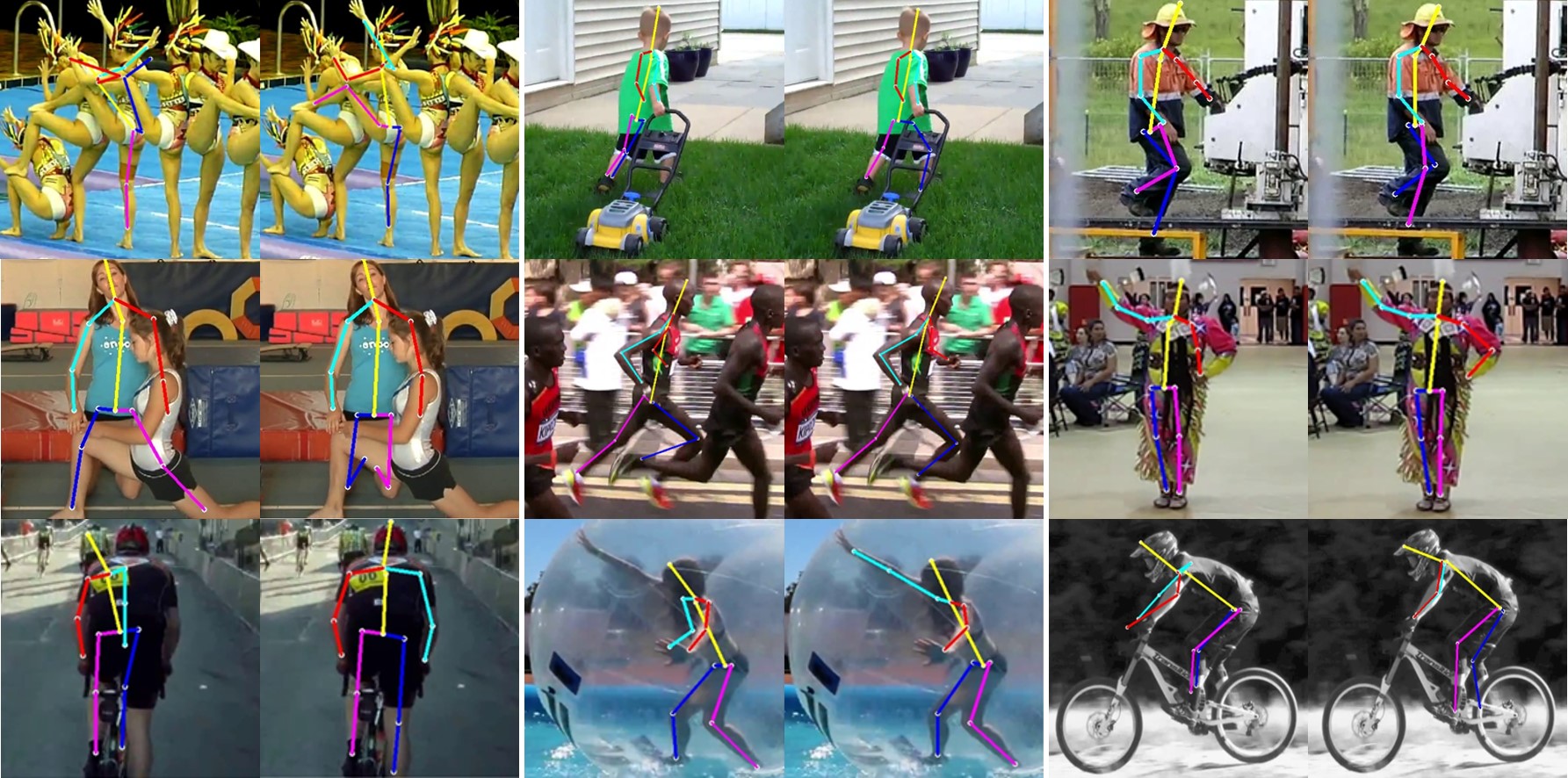}
	\caption{ Comparisons of the HRNet~\cite{sun2019deep} trained without (left side) and with (right side) our Adversarial Semantic Data Augmentation. }
	\label{fig:mpii_valid_results}
\end{figure}

\subsection{Qualitative Results}
\begin{figure}[t]
	\centering
	\includegraphics[width=1.0\columnwidth]{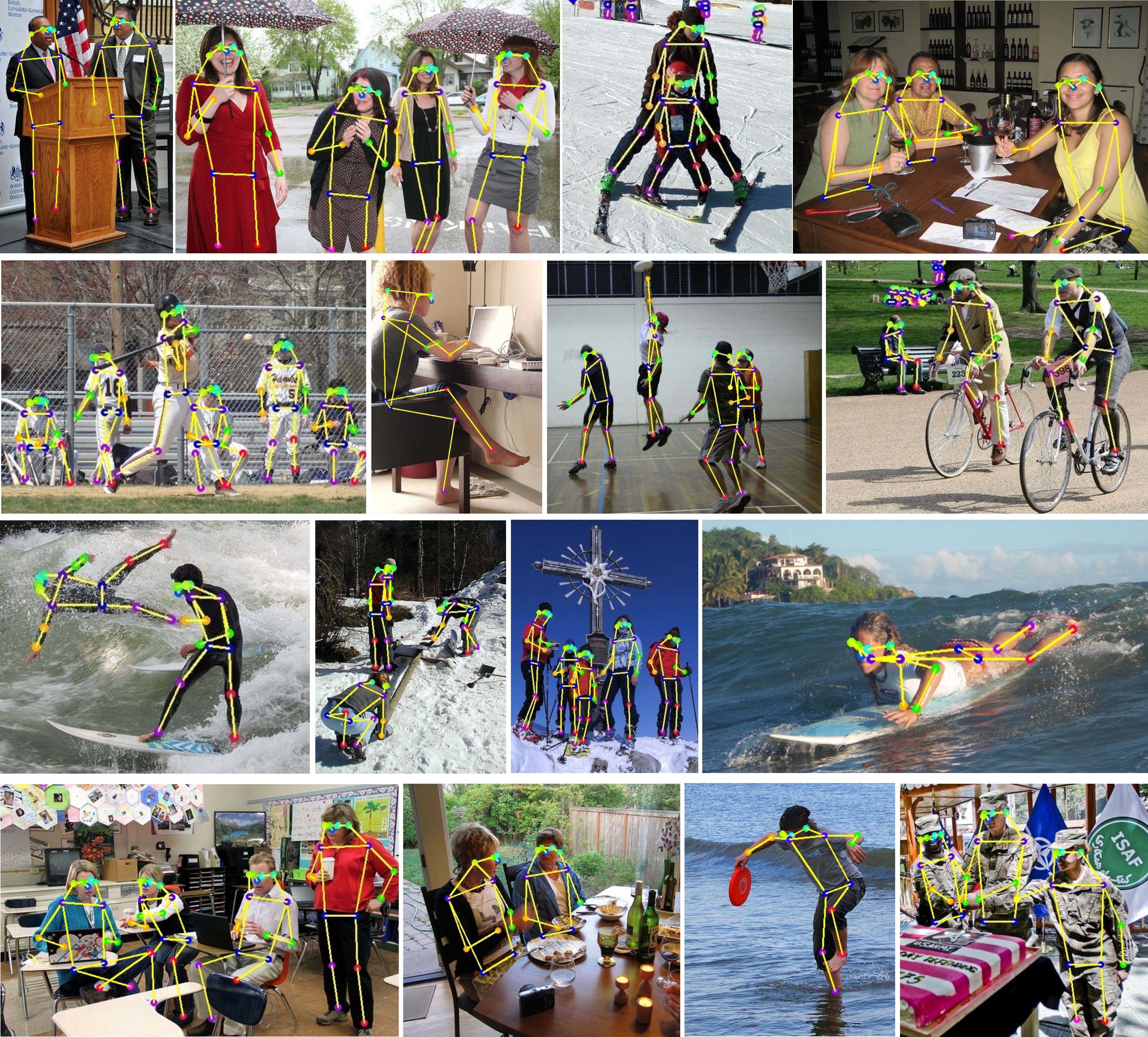}
	\caption{Examples of estimated poses on the COCO test set.}
	\label{fig:COCO_result}
\end{figure}

Figure~\ref{fig:mpii_valid_results} displays some pose estimation results obtained by HRNet without (left size) and with (right side) our ASDA. We can observe that original HRNet is confused by symmetric appearance (e.g. the left and right legs in~\{\textit{rows.1, cols. 3}\}), heavy occlusion (e.g., the right ankle in ~\{\textit{rows.1 cols. 2}\}) and nearby person (e.g., multiple similar legs and arms in~\{\textit{rows.1, cols. 1}\}). Note that image in~\{\textit{rows.1, cols. 1}\} is an extremely challenging case so that few of the keyponts are correctly predicted by the original HRNet. By generating tailored semantic augmentation for each input image, our ASDA largely improves the performance of the original HRNet in the extremely challenging cases. Figure~\ref{fig:COCO_result} shows some pose estimation results obtained by our approach on the COCO test dataset.

\subsection{Ablation Studies} \label{exp:ablation}
In this section, we conduct ablative analysis on the validation set of MPII dataset. The baseline is HRNet-W32~\cite{sun2019deep} which achieved PCKh@0.5 at 90.3\% by performing flipping and single scale in inference. During baseline training, the data augmentation adopts global spatial transformation including random rotation $(−30^{\circ}, 30^{\circ})$ ,random scale $(0.75, 1.25)$ and flipping. The results are shown in Table~\ref{table:ablation_study}~(a). 

The MPII dataset provide visibility annotations for each keypoint, which enables us to conduct ablative analysis on the  subset of invisible keypoints and study the effect of our method on improving the occlusion cases. The results are shown in Table~\ref{table:ablation_study}~(b). 

\begin{table}[h] 
	\centering
	\tabcolsep=2.0pt
	\caption{Ablation studies on the MPII validation set (PCKh@0.5)}
	\centering
	\begin{tabular}{l|ccccccc|c}
		\multicolumn{9}{c}{(a) Results evaluated on all keypoints}\\
		\hline
		Method & Hea. &	Sho. &	Elb. & Wri.	& Hip.	& Kne.	& Ank. & Total \\
		\hline
		\textbf{Baseline} & 97.1 & 95.9 & 90.3 & 86.4 & 89.1 & 87.1 & 83.3 & 90.3 \\
		\textbf{+ROR} & 97.0 & 96.2 & 90.9 & 86.9 & 89.3 & 86.9 & 82.9 & 90.5 \\
		\textbf{+SDA~~~(Ours)} & 97.2 & 96.3 & 91.2 & 86.9 & 90.0 & 87.2 & 83.7 & 90.8 \\
		\textbf{+ASDA~(Ours)} & \textbf{97.6}  & \textbf{96.6}  & \textbf{91.5}  & \textbf{87.3}  & \textbf{90.5}  & \textbf{87.5} & \textbf{84.5} & \textbf{91.2} \\
		\hline
		\multicolumn{9}{c}{}\\
		\multicolumn{9}{c}{(b) Results evaluated only on invisible keypoints}\\
		\hline
		\textbf{Baseline} & - & 90.9 & 73.6 & 61.9 & 81.8 & 71.7 & 61.8 & 74.2 \\
		\textbf{+ROR} & - & 92.0 & 74.9 & 63.2 & 82.7 & 71.6 & 61.6 & 74.9 \\
		\textbf{+SDA~~~(Ours)} & - & 91.8 & 75.1 & 63.0 & 84.1 & 71.7 & 63.3 & 75.4 \\
		\textbf{+ASDA~(Ours)} & -  & \textbf{92.7}  & \textbf{75.1}  & \textbf{65.1}  & \textbf{84.8}  & \textbf{71.8} & \textbf{63.4} & \textbf{76.1} \\
		\hline
		\multicolumn{9}{p{10cm}}{\textbf{Baseline:} The original HRNet-W32 \cite{sun2019deep}. The following experiments is all based on this baseline.} \\
		\multicolumn{9}{p{10cm}}{\textbf{+ROR:} Adopt data augmentation of Randomly Occluding and Repeating (ROR) the keypoints patch~\cite{ke2018multi} on training HRNet-W32.} \\
		\multicolumn{9}{p{10cm}}{\textbf{+SDA:} Adopt our Semantic Data Augmentation (SDA) scheme on training HRNet-W32, the augmentation parameters are adjusted randomly from a uniform distribution in the neighborhood space of $(1,0,0,0)$.} \\
		\multicolumn{9}{p{10cm}}{\textbf{+ASDA:} Adop our Adversarial Semantic Data Augmentation (ASDA) scheme on training HRNet-W32, the augmentation parameters are online adjusted by the generative network in an adversarial way.}
	\end{tabular}
	\label{table:ablation_study}
\end{table}

\textbf{With Vs. Without Semantic Data Augmentation.}
We first evaluate the effect of the Semantic Data Augmentation scheme. As shown in Table~\ref{table:ablation_study}~(a), \textbf{+SDA} outperforms the \textbf{Baseline} with a large margin by 0.5\%. 
Note that our SDA scheme consistently achieved improvements on all keypoints. Especially, our SDA achieves 0.9\%, 0.5\% and 0.4\% improvements on elbow, wrist and ankle respectly, which are considered as the most challenging keypoints to be localized.
In Table~~\ref{table:ablation_study}~(b), we can observe a more significant improvement brought by SDA.
The result demonstrate that the semantic local pixel manipulation of our SDA effectively augment training data and elevate the performance of pose estimation.

Both SDA and Randomly Occluding and Repeating (ROR) the keypoints patch~\cite{ke2018multi} augment training data by manipulate the local pixel. 
However, ROR achieves 0.3\% lower average PCKh@0.5 than our SDA. Moreover, ROR even brings negtive effects to baseline model when localizing keypoints like knee and ankle. These results demonstrate that various segmented body parts with high semantics used in our SDA play an key role for improving pose estimtion performance.

\textbf{Random Vs. Adversarial Augmentation.}
Based on the SDA scheme, we found that Adversarial SDA can further improve the accuracy by online adjusting augmentation parameters. As shown in the table~\ref{table:ablation_study}~(a), \textbf{+ASDA} consistently outperforms \textbf{+SDA} on all keypoints and achieve 0.4\%  higher average PCKh@0.5. For invisible keypoints, ASDA outperforms baseline and SDA by 1.9\% and 0.7\%  PCKh@0.5 score. As discussed in Sec. \ref{sec:adversarial_learning}, our ASDA can further improve performance due to the adversarial learning strategy which generates tailored samples for training pose estimation network.

\textbf{Sensitivity Analysis.} The part number $N$ as a hyper-parameter is configured manually. We test different $N$ values during training and the PCKh@0.5 score on the MPII validation set is shown in Table \ref{table:numParts}. Less than 3 parts, the performance maintain roughly the same. 
Begin with 4 parts, the performance sharply drop along the increasing of part number. We infer that too many parts will generate too hard training samples for pose estimation network which misleads network to learn unrealistic cases.

\begin{table}[h]
	\centering
	\tabcolsep=2.0pt
	\caption{Ablation studies of different number of body parts $N$.}
	\begin{tabular}{c|ccccccc|c}
		\hline%
		Part Num & Hea. & Sho. &	Elb. & Wri.	& Hip. & Kne. & Ank. & Total \\
		\hline
		1 & \textbf{97.6}  & 96.6  & \textbf{91.5}  & \textbf{87.3}  & 90.5  & \textbf{87.5} & \textbf{84.5} & \textbf{91.2} \\
		2 & 97.5  & 96.6  & \textbf{91.5}  & 86.9  & 90.1  & 87.4 & 83.8 & 91.0 \\
		3 & 97.3  & \textbf{96.8}  & 91.3  & 86.9  & \textbf{90.6}  & 87.4 & 83.6 & 91.0 \\
		4 & 97.4  & 96.3  & 91.1  & 86.2  & 90.3  & 87.0 & 83.6 & 90.7 \\
		6 & 97.2  & 96.2  & 90.4  & 85.2  & 90.0  & 86.0 & 82.1 & 90.1 \\
		8 & 97.0  & 95.7  & 89.3  & 83.8  & 89.3  & 85.6 & 81.4 & 89.4 \\
		\hline
	\end{tabular}
	\label{table:numParts}
\end{table}

\textbf{Apply on Different Networks.} As shown in Table \ref{table:different_model}, we report the performance of different networks trained with our ASDA. By applying our ASDA, the SOTA networks consistently achieved improvements. Especially on the challenging keypoints such as elbow, wrist, knee and ankle, our ADSA enhances the network significantly. This result exhibits the universality of our ADSA scheme.

\begin{table}[h]
	\centering
	\caption{Result of applying on different network.}
	\centering
	\begin{tabular}{l|ccccccc|cc}
		\hline
		Method& Hea. &	Sho. &	Elb. & Wri.	& Hip.	& Kne.	& Ank. & Total \\
		\hline
		2-Stacked HG & $96.6$ & $95.4$ & $89.7$ & $84.7$ & $88.7$ & $84.1$ & $80.7$ & $89.1$ \\
		2-Stacked HG+ASDA & \textbf{96.8} & \textbf{95.8} & \textbf{90.5} & \textbf{85.5} & \textbf{89.3} & \textbf{85.5} & \textbf{81.9} & \textbf{89.8} \\
		\hline
		8-Stacked HG & $96.9$ & $95.9$ & $90.8$ & $86.0$ & $89.5$ & $86.5$ & $82.9$ & $90.2$ \\
		8-Stacked HG+ASDA &\textbf{97.5} & \textbf{96.5} & \textbf{91.6} & \textbf{87.3} & \textbf{90.5} & \textbf{87.7} & \textbf{83.5} & \textbf{91.1} \\
		\hline
		SIM-ResNet50 & $96.4$ & $95.3$ & $89.0$ & $83.2$ & $88.4$ & $84.0$ & $79.6$ & $88.5$ \\
		SIM-ResNet50+ASDA & \textbf{96.8} & \textbf{95.8} & \textbf{89.7} & \textbf{83.9} & \textbf{89.5} & \textbf{85.1} & \textbf{80.5} & \textbf{89.3} \\
		\hline
		SIM-ResNet101 & $96.9$ & $\textbf{95.9}$ & $89.5$ & $84.4$ & $88.4$ & $84.5$ & $80.7$ & $89.1$ \\
		SIM-ResNet101+ASDA & \textbf{97.2} & \textbf{95.9} & \textbf{90.0} & \textbf{85.2} & \textbf{89.7} & \textbf{86.0} & \textbf{82.3} & \textbf{90.0} \\
		\hline
		HRNet-W32 & $97.1$ & $95.9$ & $90.3$ & $86.4$ & $89.1$ & $87.1$ & $83.3$ & $90.3$ \\
		HRNet-W32+ASDA & \textbf{97.6}  & \textbf{96.6}  & \textbf{91.5}  & \textbf{87.3}  & \textbf{90.5}  & \textbf{87.5} & \textbf{84.5} & \textbf{91.2} \\
		\hline
		HRNet-W48 & $97.2$ & $96.1$ & $90.8$ & $86.3$ & $89.3$ & $86.6$ & $83.1$ & $90.4$ \\
		HRNet-W48+ASDA & \textbf{97.3}  & \textbf{96.5}  & \textbf{91.7}  & \textbf{87.9}  & \textbf{90.8}  & \textbf{88.2} & \textbf{84.2} & \textbf{91.4} \\
		\hline
	\end{tabular}
	\label{table:different_model}
\end{table}
\textbf{Compare with methods that also use parsing information.}
 Nie et al~\cite{nie2018human} also use parsing information and improves the 8-stacked hourglass from 90.2\% to 91.0\% on MPII validation set.
The improvement is slightly lower than ASDA that improves the 8-stacked hourglass from 90.2\% to 91.1\%. In addition, \cite{nie2018human} uses 2-stacked hourglass as Parsing Encoder to predict the parameters of an adaptive convolution, which introduces extra parameters and computation burden. Moreover, the parsing annotation and keypoints annotation of LIP are both used in the training of Parsing Encoder while our ASDA only uses the parsing annotation.

\section{Conclusions}

In this work, we proposed Semantic Data Augmentation (SDA) which locally pasted segmented body parts with various semantic granularity to synthesize challenging cases.   
Based on the SDA, we further proposed Adversarial Semantic Data Augmentation which exploit a generative network to online adjust the augmentation parameters for each individual training image in an adversarial way. Improved results on public benchmark and comprehensive experiments have demonstrated the effectiveness of our methods. Our ASDA is general and independent on network. We hope our work can provide inspiration on how to generate tailored training samples for other tasks.

~\\
\noindent\textbf{Acknowledgement.} This work was supported by the National Natural Science Foundation of China under grant 61871435 and the Fundamental Research Funds for the Central Universities no. 2019kfyXKJC024.
\clearpage
%
%
\bibliographystyle{splncs04}
\bibliography{eccv2020submission}
\end{document}